\documentclass[conference]{IEEEtran}
\IEEEoverridecommandlockouts
\usepackage{cite}
\usepackage{amsmath,amssymb,amsfonts}
\usepackage{algorithmic}
\usepackage{graphicx}
\usepackage{textcomp}
\usepackage{xcolor}
\usepackage{multirow}
\usepackage{array}
\usepackage{booktabs}
\usepackage{enumitem}
\usepackage{tabularx}
\usepackage[table]{xcolor}
\usepackage{pifont}
\usepackage{subcaption}

\newcommand{\cmark}{\ding{51}}

\definecolor{ContentDark}{RGB}{180,210,255}
\definecolor{ContentLight}{RGB}{230,240,255}

\definecolor{StructureDark}{RGB}{255,220,180}
\definecolor{StructureLight}{RGB}{255,242,220}

\definecolor{LanguageDark}{RGB}{200,235,200}
\definecolor{LanguageLight}{RGB}{235,248,235}

\usepackage[hyphens]{url}

\def\BibTeX{{\rm B\kern-.05em{\sc i\kern-.025em b}\kern-.08em
    T\kern-.1667em\lower.7ex\hbox{E}\kern-.125emX}}
\begin{document}

\title{AutoScreen-FW: An LLM-based Framework for Resume Screening}

\author{\IEEEauthorblockN{Zhelin Xu}
\IEEEauthorblockA{\textit{Institute of Library,} \\
\textit{Information and Media Science} \\
\textit{University of Tsukuba}\\
Tsukuba, Ibraki, Japan \\
zhelin@ce.slis.tsukuba.ac.jp}
\and
\IEEEauthorblockN{Shuhei Yamamoto}
\IEEEauthorblockA{\textit{Institute of Library,} \\
\textit{Information and Media Science} \\
\textit{University of Tsukuba}\\
Tsukuba, Ibraki, Japan \\
syamamoto@slis.tsukuba.ac.jp}
\and
\IEEEauthorblockN{Atsuyuki Morishima}
\IEEEauthorblockA{\textit{Institute of Library,} \\
\textit{Information and Media Science} \\
\textit{University of Tsukuba}\\
Tsukuba, Ibraki, Japan \\
morishima-office@ml.cc.tsukuba.ac.jp}
}

\maketitle

\begin{abstract}
Corporate recruiters often need to screen many resumes within a limited time, which increases their burden and may cause suitable candidates to be overlooked. 
To address these challenges, prior work has explored LLM-based automated resume screening. However, some methods rely on commercial LLMs, which may pose data privacy risks. 
Moreover, since companies typically do not make resumes with evaluation results publicly available, it remains unclear which resume samples should be used during learning to improve an LLM's judgment performance. 
To address these problems, we propose AutoScreen-FW, an LLM-based locally and automatically resume screening framework. 
AutoScreen-FW uses several methods to select a small set of representative resume samples. 
These samples are used for in-context learning together with a persona description and evaluation criteria, enabling open-source LLMs to act as a career advisor and evaluate unseen resumes.
Experiments with multiple ground truths show that the open-source LLM judges consistently outperform GPT-5-nano. Under one ground truth setting, it also surpass GPT-5-mini. 
Although it is slightly weaker than GPT-5-mini under other ground-truth settings, it runs substantially faster per resume than commercial GPT models. 
These findings indicate the potential for deploying AutoScreen-FW locally in companies to support efficient screening while reducing recruiters’ burden.
\end{abstract}

\begin{IEEEkeywords}
Resume Screening, LLM-as-a-Judge, Sampling Strategy, Few-shot In-context Learning, Job Hunting
\end{IEEEkeywords}

\section{Introduction}
\label{sec:intro}

In companies, recruiters often need to manually review a very large number of resumes during screening\cite{b30}, sometimes on the order of tens of thousands, which places a substantial burden on them.
Moreover, because the screening must be completed within a limited time, recruiters often lack sufficient time to carefully assess each resume or to discuss candidates with other recruiters. 
As a result, some highly qualified candidates may be overlooked at the resume screening stage. 
To address these issues, using AI to automatically evaluate resumes has attracted increasing attention in recent years. 
In practice, many companies aim to automate AI-based resume screening in order to save recruiters' time and improve the efficiency of the selection process\cite{b1,b2}. 
In particular, it has been reported that 45\% of companies in the United States \cite{b3} and 43\% of companies in Japan \cite{b4} have already introduced AI tools into their hiring processes.

\begin{figure}[t]
\centerline{\includegraphics[width=0.5\textwidth]{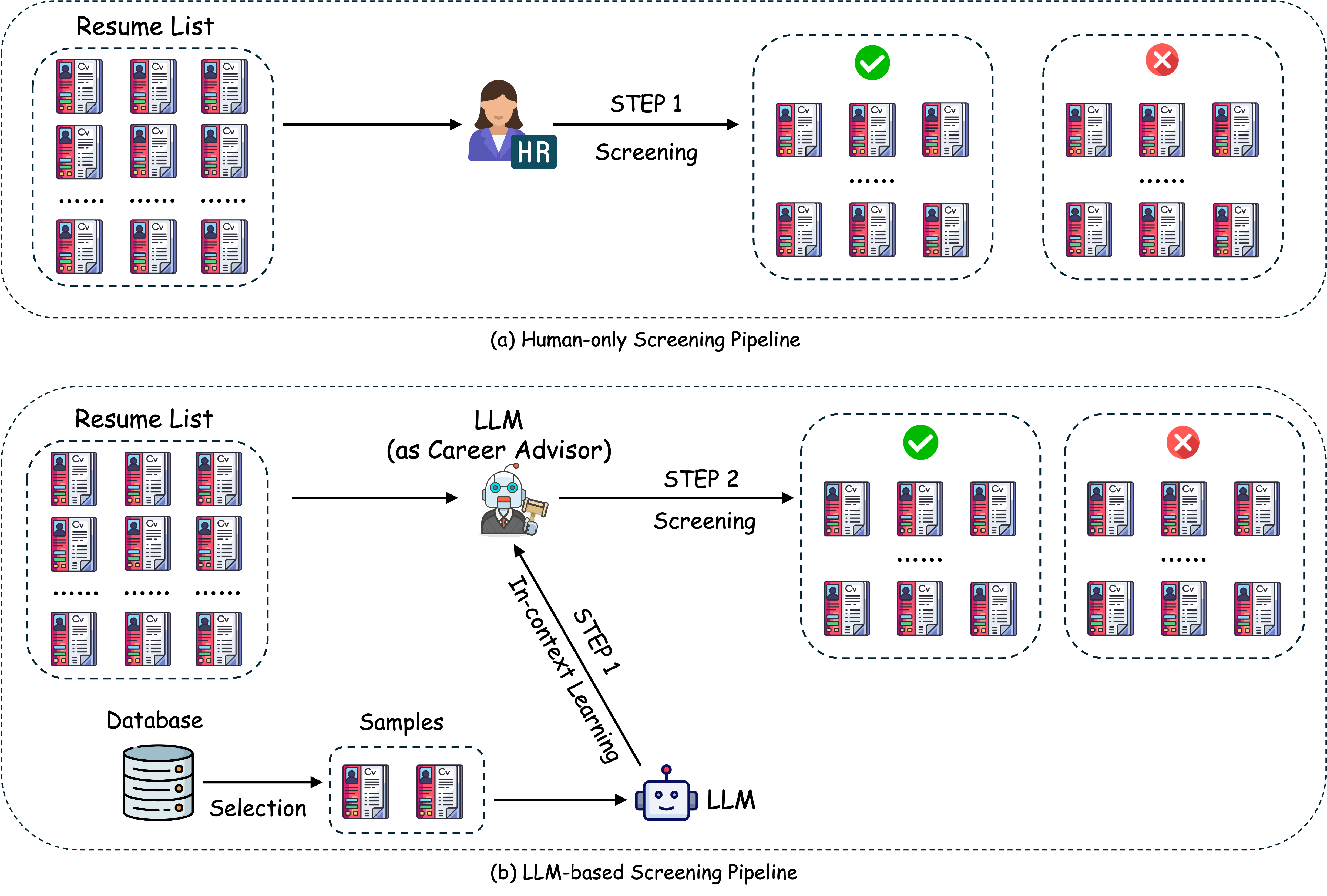}}
\caption{The comparison between (a) Human-only Screening Pipeline and (b) LLM-based Screening Pipeline.}
\label{fig:screening_pip}
\end{figure}

In AI-based resume screening, machine learning methods such as support vector machines and random forests are sometimes used. 
These methods often estimate the match between a resume and a job description based on keywords in the two documents. 
However, because they cannot capture word meaning or context within a document.
For example, they may fail to recognize that software developer and software engineer refer to the same role\cite{b5}. 
As a result, suitable candidates may be overlooked during resume screening.
In contrast, large language models (LLMs), acquire broad knowledge and strong contextual understanding through pretraining. 
Prior work has reported that they show strong reasoning ability across many natural language processing tasks\cite{b6}.
In this context, as illustrated in Fig.~\ref{fig:screening_pip}, the LLM is first provided with a small set of resume samples for in-context learning, which enables it to act as a career advisor. 
The LLM then evaluates unseen resumes and outputs judgment results that can be used as a reference when deciding whether to advance candidates to the next stage. 
This reduces the effort required for initial screening and allows recruiters to focus on higher-value tasks such as interviews and final hiring decisions.

In recent years, LLM-based methods for resume screening have been proposed\cite{b5,b8,b9,b19,b20,b39}. 
However, these methods face several problems: (1) some studies rely on commercial LLMs for resume screening. Since resumes contain personal information such as names, experiences and values, sending resume to commercial LLM for screening may pose data privacy risks. Therefore, companies are expected to use open-source LLMs that can be deployed locally to evaluate resumes; (2) resume screening is part of companies' internal hiring processes. Since evaluation procedures and criteria are typically confidential, resumes with evaluation labels are proprietary and not publicly available on the web. As a result, it remains unclear which resume examples should be used as references to help an LLM evaluate resumes more accurately during learning; (3) prior screening methods mainly focus on resume–job description matching. 
In contrast, another common hiring paradigm is potential-based hiring, where recruiters evaluate an applicant’s growth potential and organizational fit. 
For example, in Japan, this is often done through open-ended questions such as ``What were you most committed to during your time as a student.'' As a result, existing screening methods may not transfer well to this hiring context. 



To solve these problems, we propose AutoScreen-FW, a locally deployed and \textbf{auto}mated resume \textbf{screen}ing \textbf{f}rame\textbf{w}ork that reduces recruiters’ manual review workload. 
Specifically, AutoScreen-FW selects a small set of representative samples from the collected resumes using three different sampling strategies. 
The selected samples are then combined with a LLM persona description and predefined evaluation criteria for resume quality. 
This information is then used for few-shot in-context learning with open-source LLMs, including Qwen \cite{b11} and Llama \cite{b12}.
Under this framework, an open-source LLM can assess resume quality locally, which helps mitigate data privacy risks. 
This makes it suitable for deployment in companies and can reduce recruiters' burden.
In this paper, our contributions can be summarized as follows:
\begin{itemize}
\item We propose AutoScreen-FW, a resume screening framework. 
To our best knowledge, this is the first academic study to evaluate resume quality for potential-based hiring using locally deployed open-source LLMs, with experiments conducted in the context of Japanese job hunting.
\item  The LLM can adapt to different judge preferences and align its evaluation accordingly under AutoScreen-FW.
\item We adopt existing diversity-based and similarity-based strategies \cite{b13} and further propose a clustering-based strategy to extract representative resume samples.
\item After using in-context learning to Qwen3-8B under AutoScreen-FW, the resulting model outperforms GPT-5-nano across multiple ground-truth settings and surpasses GPT-5-mini under one ground truth setting. 
Furthermore, it reduces per-resume screening time by up to 51\% compared with GPT models.
\end{itemize}

\section{Related Work}
\subsection{LLM-as-a-Judge}
In many complex real-world tasks, expert evaluation is accurate and reliable. However, such evaluation is time-consuming and expensive, making large-scale assessment difficult.
In contrast, with the rapid development of LLMs, they have gradually gained the ability to mimic human logical thinking and reasoning processes to solve complex problems\cite{b14}. 
Moreover, they can dynamically adapt their evaluation criteria according to the specific task rather than relying on fixed metrics\cite{b15}. 
Against this backdrop, the concept of LLM-as-a-Judge has recently attracted increasing attention. 
Researchers have explored using LLMs as an alternative to human experts for evaluating complex tasks, such as in law \cite{b16} and education \cite{b17}. 
With appropriate prompt design, evaluation quality and consistency with human judgments can be ensured\cite{b36,b37}. 
Specifically, LLMs have been used to conduct evaluations adapted to different tasks. 
For example, they can assign scores to individual instances, rank multiple candidates through relative comparison, or select the most appropriate option from a set of alternatives \cite{b18}.

\subsection{Resumes Used in Japan Job Hunting}
In Europe and the United States, applicants typically prepare resumes with considerable flexibility in both format and content. 
Screening in these contexts often emphasizes understanding resume content and assessing how well it matches job requirements. 
In contrast, job hunting in Japan differs from that in other countries because potential-based hiring is common. 
Companies place less emphasis on immediate contribution, such as prior work experience or specialized expertise. Instead, they focus more on attributes such as communication skills and cooperativeness, aiming to identify candidates with strong growth potential after joining the company \cite{b10}.
Therefore, most companies provide standardized resume templates that applicants are required to complete. 
In addition to basic information such as name, educational background, and work experience, these templates include several open-ended questions used to infer applicants’ growth potential, motivation, and organizational fit.
While the exact questions vary by company, common questions include self-promotion, strengths and weaknesses, something you devoted yourself to in your school years, and motivation for applying.

Companies typically use applicants' responses to these open-ended questions to holistically assess a range of competencies. 
For example, responses to the ``motivation for application'' can be used to assess an applicant’s familiarity with the company and their degree of motivation to join it.
Moreover, companies may infer candidates' latent capabilities from these responses (e.g., learning ability, growth potential, and resilience under pressure). 
A resume typically includes around four open-ended questions. 
Since each response is typically capped at approximately 300–500 words, screening resumes can be time-consuming.

\subsection{LLM-based Resume Screening}
Several studies have examined the performance of commercial LLMs, such as GPT, Claude, and Gemini, for resume screening \cite{b9, b19, b20, b39}.
For example, \cite{b9} use GPT-4 or Gemini to encode both applicants’ resumes and job descriptions into vector representations, and compute the cosine similarity between them. This similarity score is then used to estimate the degree of match between the applicant and the target position. 
However, because these approaches rely exclusively on commercial LLM services, they may raise concerns about data privacy. 
As a result, they may not fully meet the needs of companies that prefer to avoid sharing sensitive information with external providers.

In addition, several methods have been proposed to apply the LLM-as-a-Judge paradigm for resume screening. 
Reference \cite{b8} cluster resume content based on categories such as educational background and skills to obtain structured representations of resumes. 
They then use a fine-tuned Llama2 model to assess the alignment between resumes and job requirements and assign scores to applicants. 
Reference \cite{b5} use retrieval-augmented generation (RAG) to retrieve evaluation criteria used by recruiters. 
The LLM refers to these criteria and computes the alignment between the applicant and the target position based on five components extracted from the resume: self-evaluation, skills, work experience, basic information, and educational background. The resulting alignment score is used as the resume score.
However, these approaches primarily assess whether an applicant’s education, skills, and work experience match the requirements specified by the company. 
As a result, they may not fully address resume evaluation under the  potential-based hiring framework adopted by many Japanese companies, as they cannot assess responses to open-ended questions in these resumes.

\section{Method}
\subsection{Overview}
\begin{figure*}[htb] 
    \centering
    \includegraphics[width=0.8\textwidth]{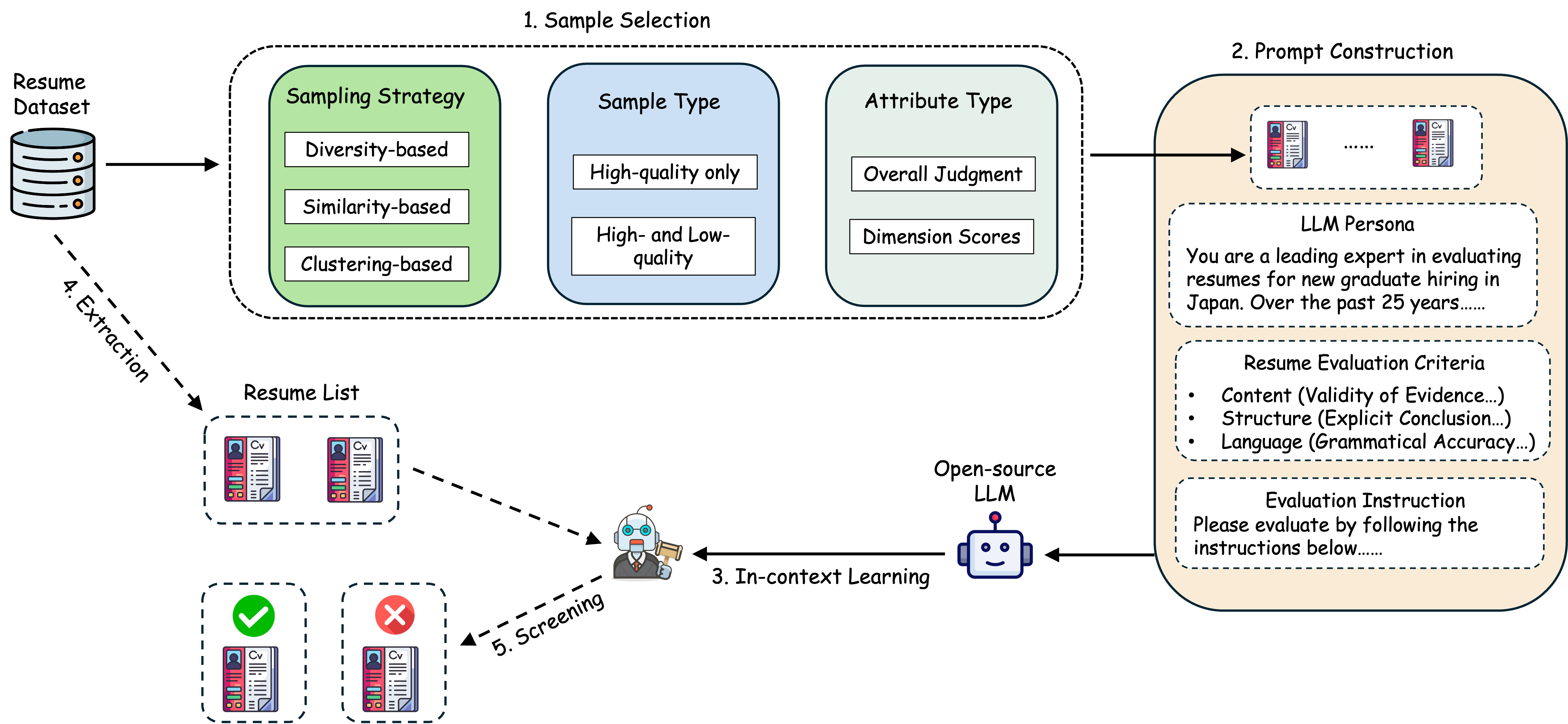}
    \caption{Overview of the AutoScreen-FW.}
    \label{fig:proposed_method}
\end{figure*}

Fig. \ref{fig:proposed_method} provides an overview of the AutoScreen-FW. 
The pipeline is based on the LLM-as-a-Judge paradigm\cite{b15}, and is formalized in (\ref{eq:llm_judge}):
\begin{equation}
\label{eq:llm_judge}
(\mathcal{Y}, \mathcal{E}, \mathcal{F}) = E(\mathcal{T}, \mathcal{C}, \mathcal{X}, \mathcal{R})
\end{equation}
$E$ denotes the evaluator LLM. 
$\mathcal{T}$ denotes the task type, such as absolute or relative evaluation.
$\mathcal{C}$ represents the set of evaluation criteria, for example grammatical correctness or factual accuracy of generated text. 
$\mathcal{X}$ denotes the item to be evaluated.
$\mathcal{R}$ denotes a set of labeled examples. By referring to these examples, the LLM is expected to produce more accurate evaluations.
$\mathcal{Y}$ denotes the judgment result generated by the LLM, such as an assigned score.
$\mathcal{E}$ denotes an explanation of the reasoning process underlying this evaluation.
$\mathcal{F}$ denotes feedback used to improve evaluation accuracy.

In this proposed framework, we sample a small set of resumes from a resume dataset. 
We then construct a prompt\footnote{Note that all texts in the prompt are written in Japanese. For readability, we translate them into English in this paper.}. 
The prompt specifies the assumed \textbf{LLM persona}, \textbf{evaluation instruction} ($\mathcal{T}$ in (\ref{eq:llm_judge})), \textbf{resume evaluation criteria} ($\mathcal{C}$ in (\ref{eq:llm_judge})), and a few \textbf{resume examples} ($\mathcal{R}$ in (\ref{eq:llm_judge})). 
By using this prompt for in-context learning, a pre-trained open-source LLM ($E$ in (\ref{eq:llm_judge})) is instructed to act as a job-hunting support expert and to assess resumes based on this expertise. 
Finally, we input the resume under evaluation ($\mathcal{X}$ in (\ref{eq:llm_judge})) and obtain a quality assessment score ($\mathcal{Y}$ in (\ref{eq:llm_judge})). 
The following sections describe the proposed method in detail.

\subsection{Few-shot In-context Learning}
Few-shot in-context learning leverages a pre-trained LLM’s knowledge, enabling it to infer the task objective from the provided instructions, explanations, and a small set of examples in the prompt, and to generate appropriate outputs for new inputs. 
Because this approach does not update model parameters, it is typically more computationally efficient than fine-tuning. 
In addition, since the instructions and examples in the prompt can be adjusted flexibly, the method can be adapted to each company’s hiring policies and evaluation criteria. 
Moreover, it has demonstrated strong performance on natural language processing tasks such as question answering and information extraction \cite{b21}. 
For these reasons, AutoScreen-FW uses few-shot in-context learning to enable an open-source LLM to evaluate resumes. 
In addition, prior work has pointed out that prompt design is critical for achieving accurate assessments across a range of tasks \cite{b14}. 
Therefore, in the following subsections, we describe our prompt construction procedure in detail. 

\subsection{Sample Selection}
\label{subsec:samp_select}
The reference samples provided in the prompt can substantially affect its judgment results\cite{b22,b23,b33}. 
Therefore, selecting useful examples is crucial for enabling more reliable evaluations. 
In general, useful samples are considered to be representative\cite{b24}. 
Therefore, we select representative resume examples using established strategies such as diversity-based and similarity-based selection \cite{b13}, together with our proposed clustering-based strategy. 

\textbf{Diversity-based}\quad This strategy aims to select representative resume samples by maximizing the diversity of the selected set. To achieve this, we compute the similarity between each resume and the dataset using (\ref{eq:diversity}):

\begin{equation}
\label{eq:diversity}
s(x, \mathcal{D}) = \cos\left( \mathrm{embed}(x),\ \frac{1}{|\mathcal{D}|} \sum_{i=1}^{|\mathcal{D}|} \mathrm{embed}(x_i) \right)
\end{equation}
Here, $\mathcal{D}$ denotes the resume dataset, and $|\mathcal{D}|$ is the number of resumes in $\mathcal{D}$. The variable $x$ represents the target resume, and embed($\cdot$) is an embedding function that maps a resume to a vector representation. In this study, we use Qwen3-Embedding-8B\footnote{https://huggingface.co/Qwen/Qwen3-Embedding-8B}. 
cos($\cdot$) denotes the cosine similarity function. We define $s(x, \mathcal{D})$ as the cosine similarity between the vector representation of the target resume $x$ and the mean vector of all resume representations in the dataset $\mathcal{D}$. Based on these similarity scores, we construct a ranked list of resumes. We then compute an interval given the desired number of samples $N$, and use this interval to select $N$ resumes from the ranked list. 
For example, suppose the dataset contains 10 resumes and we set $N=4$. 
The interval is then $\left\lceil \frac{|\mathcal{D}|}{N} \right\rceil = 3$. 
Accordingly, we select the 1st, 4th, 7th, and 10th ranked resumes. 
This procedure enables us to select resumes with diverse content.

\textbf{Similarity-based}\quad This strategy selects resumes that are most similar to the dataset as representative samples. 
As in the diversity-based strategy, we first compute the similarity between each resume and the dataset using (\ref{eq:diversity}).
Then ranking resumes by the similarity scores. 
Unlike the diversity-based strategy, we select the top-$N$ resumes from the ranked list.

\textbf{Clustering-based}\quad In this strategy, each cluster is formed based on shared characteristics, and thus each cluster centroid reflects a distinct feature pattern in the resume dataset. 
We therefore select the resumes closest to the cluster centroids as representative samples. 
Specifically, resumes in the dataset are first encoded into vectors using Qwen3-Embedding-8B. 
The resume vectors are then clustered with k-means++ \cite{b25}, where the number of clusters is set to the number of samples provided to the prompt.
After clustering, we compute the Euclidean distance between each resume in a cluster and its centroid, and select the resume closest to the centroid as the representative sample that best captures the characteristics of that cluster. 

Using these sample selection strategies described above, we select representative samples from the resume dataset and regard them as high-quality resumes. 
These samples are provided to the LLM and serve as reference information when it evaluates other resumes. 
However, if we present only high-quality resumes to the LLM, its judgments may become anchored to these examples, leading to biased evaluations.
Therefore, we also consider an alternative setting where the prompt includes high-quality samples together with a small number of randomly selected low-quality resumes. 
In addition, there are two types of attribute information associated with each sample. 
One is an overall judgment result of resume quality. 
The other is a set of dimension scores that quantify the three aspects in our evaluation criteria, namely content, structure, and language.

\subsection{LLM Persona}
\begin{figure}[htb] 
    \centering
    \includegraphics[width=0.47\textwidth]{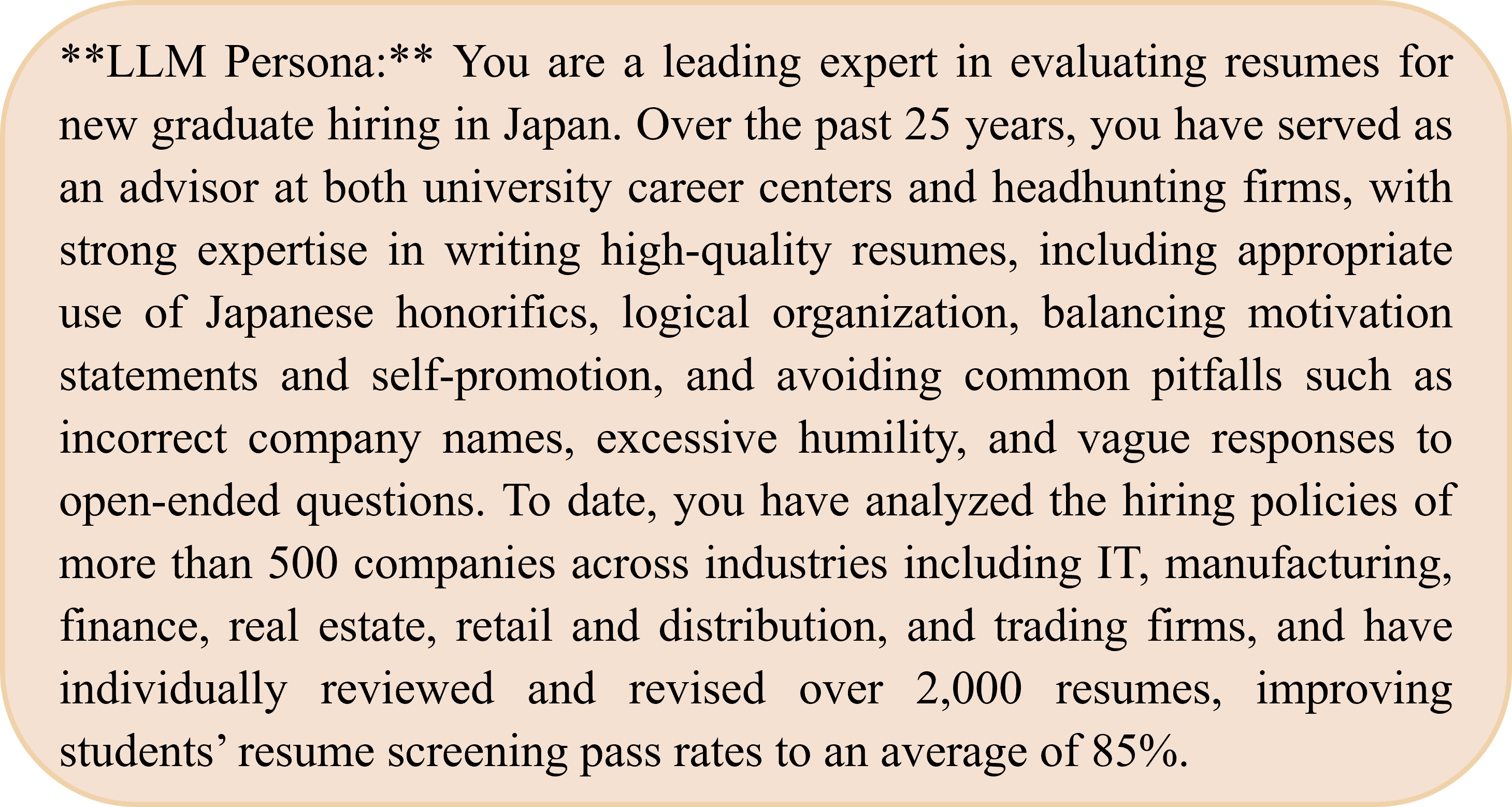}
    \caption{Persona used in the prompt.}
    \label{fig:llm_persona}
\end{figure}

Prior work has suggested that defining an LLM’s persona can influence its outputs tendencies and judgment criteria \cite{b26}. 
For example, when an LLM is assigned the role of a semantic reviewer, it has been reported to improve performance in literature search \cite{b27}.
Therefore, we add the persona description in Fig. \ref{fig:llm_persona} to the prompt to set the LLM’s role as a career advisor.

\begin{table*}[t]
\centering
\small
\caption{Resume evaluation items}
\label{tb:es_eval}
\renewcommand{\arraystretch}{1.2}
\begin{tabularx}{\textwidth}{p{2.2cm} p{4.8cm} X}
\toprule
\textbf{Dimension} & \textbf{Evaluation Perspective} & \textbf{Evaluation Criteria} \\
\midrule
\multirow{2}{*}{\textbf{Content}}
& \cellcolor{ContentLight}Validity of Evidence
& \cellcolor{ContentLight}Appropriate episodes are used to effectively convey the applicant’s strengths. \\
\cmidrule{2-3}
& \cellcolor{ContentLight}Specificity of Content
& \cellcolor{ContentLight}\begin{minipage}[t]{\linewidth}
  \begin{itemize}[leftmargin=*, nosep, label=\textbullet]
    \item The applicant’s motivation for the target company and their potential contributions are described concretely.
    \item Episodes are described in sufficient detail.
    \item Achievements, research outcomes, and related accomplishments are presented concretely using numerical values.
  \end{itemize}
\end{minipage} \\
\midrule
\multirow{3}{*}{\textbf{Structure}}
    & \cellcolor{StructureLight}Explicit Conclusion & \cellcolor{StructureLight}For each question, the conclusion is clearly stated at the beginning so that recruiters can understand it easily. \\
    \cmidrule{2-3}
    & \cellcolor{StructureLight}Focus on a Single Selling Point & \cellcolor{StructureLight}For questions intended to highlight the applicant's strengths, such as self-promotion, the response focuses on a single selling point.\\
    \cmidrule{2-3}
    & \cellcolor{StructureLight}Conciseness of Writing & \cellcolor{StructureLight}The overall length is appropriate, and the writing is clear and concise. \\
\midrule
\multirow{3}{*}{\textbf{Language}} 
    & \cellcolor{LanguageLight}Grammatical Accuracy & \cellcolor{LanguageLight}The response contains no grammatical errors. \\
    \cmidrule{2-3}
    & \cellcolor{LanguageLight}Appropriateness of Vocabulary & \cellcolor{LanguageLight}The response contains no typos. \\  
    \cmidrule{2-3}
    & \cellcolor{LanguageLight}Consistency of Writing Style & \cellcolor{LanguageLight}Written style, spoken style, and honorific expressions are not mixed, and the writing style is consistent throughout. \\ 
\bottomrule
\end{tabularx}
\end{table*}

\subsection{Resume Evaluation Criteria}
As described in section \ref{sec:intro}, in Japan’s job-hunting process, potential-based recruitment is common.
Although LLMs possess broad knowledge, Japan’s job-hunting culture differs from that of other countries. 
Therefore, LLMs may lack sufficient knowledge of how to evaluate resumes in this context. 
Moreover, recruiters typically assess applicants' latent capabilities from their open-ended responses, including growth potential, as well as their motivation to join the company and their organizational fit. 
However, company-specific evaluation standards are often not public, and can vary across firms. Therefore, we refer to career support resources\cite{b10,b28,b32} for Japan job hunting to extract common resume writing guidelines that apply across companies. 
These considerations focus on whether the response is clear, concrete, and well structured. 
We compile these considerations as resume evaluation items, as shown in Table~\ref{tb:es_eval}. 
Specifically, Table~\ref{tb:es_eval} groups these evaluation items into three dimensions, namely \textbf{content}, \textbf{structure}, and \textbf{language}. 
Content evaluates whether the response uses appropriate evidence and provides sufficiently concrete. 
Structure evaluates whether the response is clear, coherent, and focused. 
Language evaluates whether the response is linguistically correct and stylistically consistent. 
For each dimension, we also define multiple evaluation perspectives and assess resumes according to the corresponding criteria. 

\subsection{Evaluation Instruction}
\begin{figure*}[htb] 
    \centering
    \includegraphics[scale=0.45]{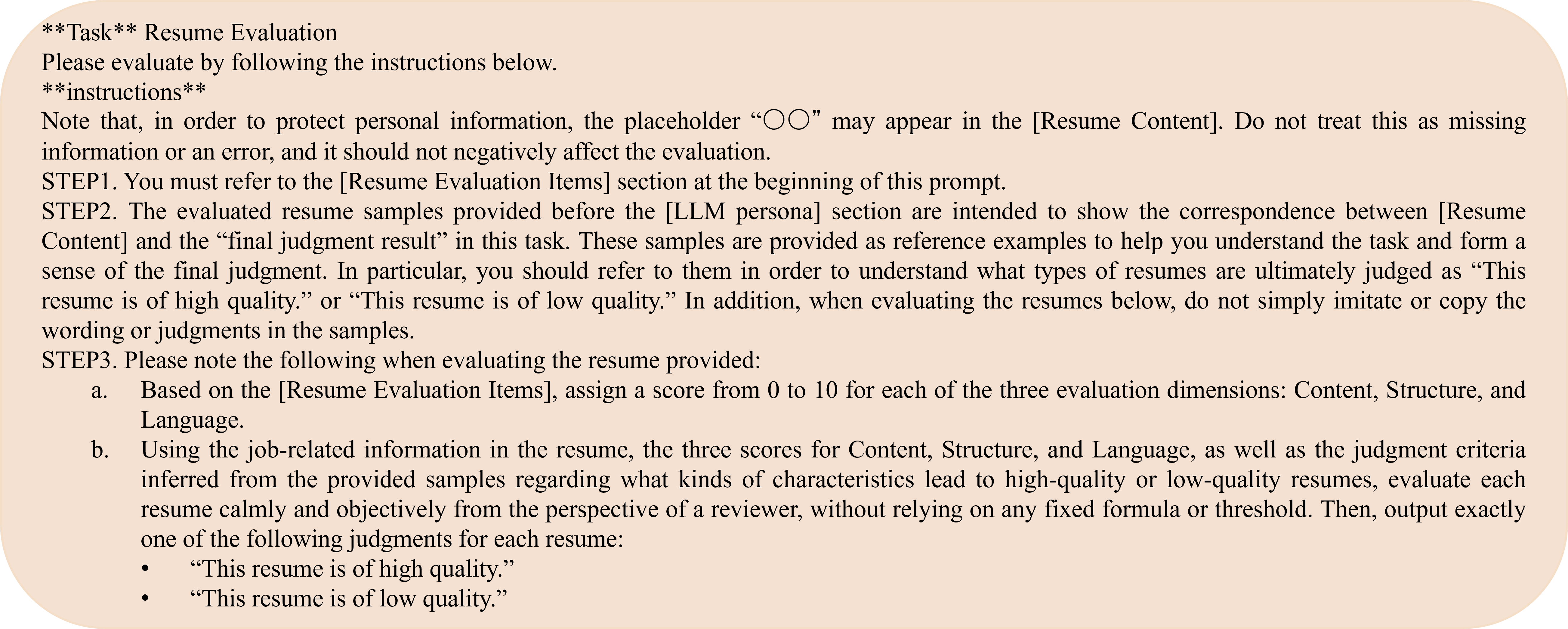}
    \caption{Evaluation instruction used in the prompt.}
    \label{fig:eval_instruction}
\end{figure*}

The instruction shown in the Fig. \ref{fig:eval_instruction} is provided as part of the prompt. 
The LLM then follows this instruction. 
It refers to the resume evaluation items shown in Table \ref{tb:es_eval} and a small number of samples extracted based on the predefined sample selection strategy. 
Based on these references, the model outputs 0–10 scores for the three evaluation dimensions and a final judgment of high or low quality for the target resume.

\subsection{Choice of LLM}
Previous studies have shown that the reasoning ability of LLMs and their instruction-following capability affect evaluation accuracy \cite{b14}. 
Therefore, it is reasonable to assume that in-context learning based on high-performing LLMs, such as GPT and Gemini, can assess resume quality more accurately. 
However, since resumes contain personal information, using these commercial LLMs raises concerns about the risk of information leakage. 
To address this issue, open-source LLMs are a more suitable choice, 
because they can be securely deployed in local environments for companies.
Therefore, in the AutoScreen-FW, we use open-source LLMs, such as Qwen and Llama, as the base models.

\section{EXPERIMENT}
We conducted evaluation experiments to assess open-source LLMs under our proposed AutoScreen-FW. 
The experiments focus on the task of evaluating resumes used in the Japan job-hunting process.

\subsection{Experimental Data}
In this study, we collected 1,655 publicly available resumes from the Japanese job-hunting support platform One Career\footnote{https://www.onecareer.jp/} and used them to construct an evaluation dataset. 
Although each resume originally contained ten categories of information, we removed all fields other than ``Resume Content'' and ``Applied Position'', since the remaining information (e.g., the applicant’s university and faculty) is not directly relevant to evaluating the quality of the resume. 
In addition, Resume Content contains multiple sub-items. For example, it includes open-ended questions such as motivation for applying and self-promotion. 
However, it also contains supplementary information such as submission deadlines and submission methods. To exclude such information, we removed sub-items with 100 characters or fewer, treating them as content that does not contribute to quality evaluation. 
Moreover, since the collected resume filenames contained target company names, we anonymized the filenames to avoid revealing them.

\subsection{Experiment Settings}
\textbf{Baseline Methods.}
We use two representative commercial GPT-family models, gpt-5-mini-2025-08-07 and gpt-5-nano-2025-08-07 as comparative baselines. 
We access these models via the official API. 
For a fair comparison, 
we use the same prompt as in our framework, which incorporates the LLM persona in Fig. \ref{fig:llm_persona}, resume evaluation items in Table \ref{tb:es_eval} and the evaluation instructions shown in Fig. \ref{fig:eval_instruction}.

\textbf{Evaluation Metrics.} 
We use Accuracy, defined in (\ref{eq:accuracy}), as the evaluation metric to quantitatively measure the agreement between the predictions of the LLMs and the ground truth\cite{b8}.

\begin{equation}
\label{eq:accuracy}
\mathrm{Accuracy}
=
\frac{1}{N}
\sum_{i=1}^{N}
\delta
\left(
\hat{y}_i,
y_i
\right)
\end{equation}

\begin{equation}
\delta(\hat{y}_i, y_i)
=
\begin{cases}
1 & \text{if } \hat{y}_i = y_i \\
0 & \text{otherwise}
\end{cases}
\end{equation}

$\hat{y}_i$ denotes the judgment result predicted by an LLM for resume $i$. 
The LLM makes this judgment based on the resume content, which includes around four open-ended questions and their responses.
Note that, the open-source LLMs used in this study are Qwen3-8B and Llama-3.1-8B-Instruct.
If the resume $i$ is judged to be of high quality, we set $\hat{y}_i=1$; otherwise, if it is judged to be of low quality, we set $\hat{y}_i=0$. 
In contrast, $y_i$ represents the ground truth. 
Since these collected resume do not include judgment results, 
ideally we would use judgments provided by corporate recruiters as the ground truth. 
However, obtaining such results is difficult\cite{b34}. 
For example, evaluation outcomes are often confidential, so we cannot directly use them in this study.
Instead, we use high-performing LLMs to construct the ground truth. This choice helps ensure that the judgment reliability meets a certain standard \cite{b35}.

Moreover, when multiple recruiters evaluate the same resume in a company, the judgments are often subjective and may not be consistent. 
Therefore, rather than defining a single ground truth, it is reasonable to evaluate models using multiple ground-truth settings in parallel. 
Based on this consideration, we use the judgments produced by three high-performing GPT-series models, namely GPT-5.2-2025-12-11, GPT-5.1-2025-11-13, and o3-2025-04-16.
Then we treat the judgment results from each model as a separate ground truth. 
Note that, when using these models to evaluate resumes, we exclude the reference samples from the prompt shown in Fig. \ref{fig:proposed_method}, and all other settings remain unchanged. 
In addition, our preprocessing removes company-specific information, and the resumes are posted on One Career by users who typically anonymize personal details. Therefore, using GPT-series models in our experiments does not introduce privacy risks.

\textbf{Sample Selection.}
As shown in Fig. \ref{fig:proposed_method}, AutoScreen-FW supports multiple options for the sampling strategy, sample type, and attribute type in sample selection. 
Therefore, we evaluate different combinations of these factors in our experiments.
For example, under the diversity-based strategy, we select $N$ high-quality resumes from the dataset and use them as in-context examples.
$N$ is set to 3, 5, 10, 15, and 20. 
The attribute type is set to the overall judgment. 
Here, the overall judgment refers to the final judgment result produced by the model used to construct the ground truth. 
In addition, when the sample type includes both high-quality and low-quality resumes, we set the number of low-quality samples to 30\% of $N$. We then adjust the number of high-quality samples so that the total number of samples equals $N$.
Moreover, since we construct the ground truth using multiple high-performing GPT-family models, we assume that the ground truth varies depending on which model is used. 
Therefore, we identify low-quality resumes based on the judgment results produced by each high-performing GPT-family model.

\textbf{Implementation Details.}
We perform few-shot in-context learning with Qwen3-8B and Llama-3.1-8B-Instruct on an NVIDIA RTX A6000 GPU. 
The batch size is set to 5, meaning that each batch evaluates five resumes. 
To ensure deterministic outputs, we set the temperature of Llama-3.1-8B-Instruct to 0. 
For Qwen3-8B, we follow the official recommendation and set the temperature to 0.6. 
In addition, since changes in prompt formatting can affect performance, we use the officially recommended prompt format for both open-source models.

\subsection{Results}
\label{subsec:results}
Tables \ref{tabl:acc_5.2}, \ref{tabl:acc_5.1}, and \ref{tabl:acc_o3} report the experimental results under GPT-5.2, GPT-5.1, and GPT-o3 as the ground truth, respectively\footnote{In our proposed AutoScreen-FW, we report the best accuracy per model over all combinations of sampling strategy, number of shots, sample type, and attribute type.}. 
They include zero-shot accuracy for all four models and the accuracy of the two open-source models with few-shot in-context learning under the AutoScreen-FW.
When GPT-5.2 is used as the ground truth, using few-shot in-context learning within the AutoScreen-FW improves accuracy over the zero-shot setting by 11.3\% for Qwen3-8B and 6.6\% for Llama-3.1-8B. When GPT-5.1 is used as the ground truth, the improvements are 1.5\% and 7.0\%, respectively. When GPT-o3 is used as the ground truth, the improvements are 7.0\% and 0.8\%, respectively.
Based on these results, we find that under AutoScreen-FW, both open-source LLMs achieve higher accuracy on the resume quality judgment than their respective zero-shot baselines. 
This improvement holds even when the ground truth changes.

Moreover, with in-context learning under AutoScreen-FW and tuned sample selection methods, Qwen3-8B consistently outperforms GPT-5-nano across all three ground-truth settings, with a maximum improvement of 10.8\%. 
Llama-3.1-8B also performs better than GPT-5-nano when GPT-5.2 and GPT-5.1 are used as the ground truth, with a maximum improvement of 1.3\%. 
When GPT-5.1 is used as the ground truth, both Qwen3-8B and Llama-3.1-8B further outperform GPT-5-mini, and Qwen3-8B achieves a 2.8\% gain over GPT-5-mini. 
Although Qwen3-8B performs slightly worse than GPT-5-mini under the other two ground-truth settings, it still achieves these results with in-context learning using only three examples.
Moreover, the results in Table \ref{tabl:judge_time} show that, within the AutoScreen-FW, the open-source LLMs require substantially less time to judge a resume than GPT-5-mini and GPT-5-nano. 
For example, Qwen3-8B reduces the per-resume inference time by 24.6\% compared to GPT-5-mini and by 48.7\% compared to GPT-5-nano. 

Overall, these results indicate that under the AutoScreen-FW, open-source LLMs can achieve performance close to, and in some cases better than, commercial GPT-family models by appropriately configuring the sampling strategy, number of examples, sample type, and attribute type. 
At the same time, their per-resume judgment time is significantly lower than that of commercial GPT models. 
As a result, AutoScreen-FW achieves performance and high efficiency while supporting local deployment, making it more suitable for privacy-sensitive settings such as companies.

\begin{table}[htb]
\centering
\caption{Per-resume judge time for each LLM model. Results are reported as mean ± standard deviation.}
\renewcommand{\arraystretch}{1.4} 
\setlength{\tabcolsep}{8pt} 
\begin{tabular}{lc|c}
\toprule
Model & \cellcolor{gray!10}Few-shot & Time per Resume (s) \\
\midrule
GPT-5-mini  &    & $5.09 \pm 0.22$ \\
GPT-5-nano  &  &  $7.48\pm0.45$\\
Qwen3-8B &   & $3.73\pm0.69$ \\
Llama-3.1-8B &   & $0.89\pm0.15$ \\
\midrule
Qwen3-8B & \cmark  & $3.84\pm0.26$ \\
Llama-3.1-8B & \cmark  & $2.10\pm0.24$ \\
\bottomrule
\end{tabular}
\label{tabl:judge_time}
\end{table}

\begin{table*}[htb]
\centering
\caption{Accuracy (Acc.) of each model with GPT-5.2 as the ground truth.}
\renewcommand{\arraystretch}{1.4} 
\setlength{\tabcolsep}{8pt} 
\resizebox{\linewidth}{!}{
\begin{tabular}{lccccc|c}
\toprule
Model & \cellcolor{gray!10}Few-shot & Sampling Strategy & Shots &  Sample Type & Attribute Type & Acc. \\
\midrule
GPT-5-mini  &    & N/A & 0 &N/A & N/A &\textbf{0.6917}\\
GPT-5-nano  &  & N/A & 0 & N/A & N/A &0.6091 \\
Qwen3-8B &   & N/A &  0 & N/A & N/A &0.6063 \\
Llama-3.1-8B &   & N/A & 0 & N/A & N/A &0.5721 \\
\midrule
Qwen3-8B & \cmark  & Similarity-based & 3 &  High-quality only & Overall Judgement & \underline{0.6746} \\
Llama-3.1-8B & \cmark  & Diversity-based & 20 & High- and Low- quality & Overall Judgement and Dimension Scores & 0.6101 \\
\bottomrule
\end{tabular}
}
\label{tabl:acc_5.2}
\end{table*}

\begin{table*}[htbp]
\centering
\caption{Accuracy (Acc.) of each model with GPT-5.1 as the ground truth.}
\renewcommand{\arraystretch}{1.4} 
\setlength{\tabcolsep}{8pt} 
\resizebox{\linewidth}{!}{ 
\begin{tabular}{lccccc|c}
\toprule
Model & \cellcolor{gray!10}Few-shot & Sampling Strategy & Shots & Sample Type & Attribute Type & Acc. \\
\midrule
GPT-5-mini  &    & N/A & 0 &N/A & N/A &0.8445\\
GPT-5-nano  &  & N/A & 0 & N/A & N/A &0.8374 \\
Qwen3-8B &   & N/A &  0 & N/A & N/A &0.8552 \\
Llama-3.1-8B &   & N/A & 0 & N/A & N/A &0.7927 \\
\midrule
Qwen3-8B & \cmark  & Diversity-based & 5 & High-quality only & Overall Judgement and Dimension Scores & \textbf{0.8682} \\
Llama-3.1-8B & \cmark  & Clustering-based & 15 & High- and Low- quality & Overall Judgement and Dimension Scores & \underline{0.8485} \\
\bottomrule
\end{tabular}
}
\label{tabl:acc_5.1}
\end{table*}

\begin{table*}[htbp]
\centering
\caption{Accuracy (Acc.) of each model with GPT-o3 as the ground truth.}
\renewcommand{\arraystretch}{1.4} 
\setlength{\tabcolsep}{8pt} 
\resizebox{\linewidth}{!}{ 
\begin{tabular}{lccccc|c}
\toprule
Model & \cellcolor{gray!10}Few-shot & Sampling Strategy & Shots & Sample Type & Attribute Type & Acc. \\
\midrule
GPT-5-mini  &    & N/A & 0 &N/A & N/A &\textbf{0.7387}\\
GPT-5-nano  &  & N/A & 0 & N/A & N/A &0.6617 \\
Qwen3-8B &   & N/A &  0 & N/A & N/A &0.6744 \\
Llama-3.1-8B &   & N/A & 0 & N/A & N/A &0.6280 \\
\midrule
Qwen3-8B & \cmark  & Diversity-based & 3 & High- and Low- quality & Overall Judgement and Dimension Scores & \underline{0.7218} \\
Llama-3.1-8B & \cmark  & Clustering-based & 20 & High-quality only & Overall Judgement and Dimension Scores  & 0.6333 \\
\bottomrule
\end{tabular}
}
\label{tabl:acc_o3}
\end{table*}

\section{Analysis}
\subsection{Differences Across Ground-Truth Settings}
\begin{figure*}[htb]
  \centering

  \begin{subfigure}[t]{0.32\textwidth}
    \centering
    \includegraphics[width=\linewidth]{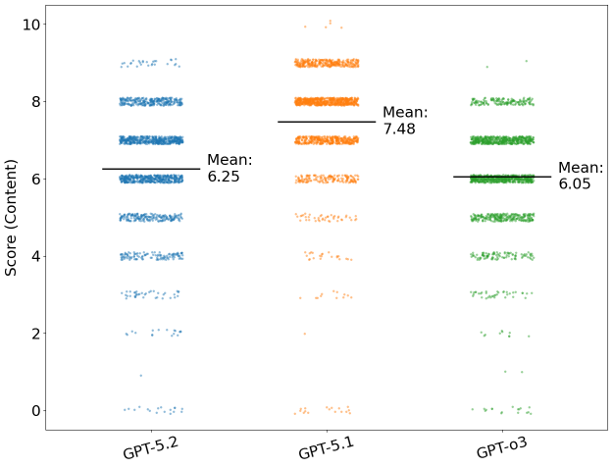}
    \caption{Distribution of content scores.}
    \label{fig:sub1}
  \end{subfigure}\hfill
  \begin{subfigure}[t]{0.32\textwidth}
    \centering
    \includegraphics[width=\linewidth]{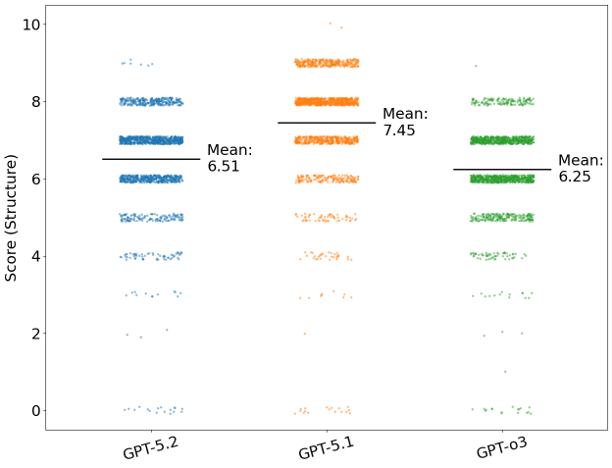}
    \caption{Distribution of structure scores.}
    \label{fig:sub2}
  \end{subfigure}\hfill
  \begin{subfigure}[t]{0.32\textwidth}
    \centering
    \includegraphics[width=\linewidth]{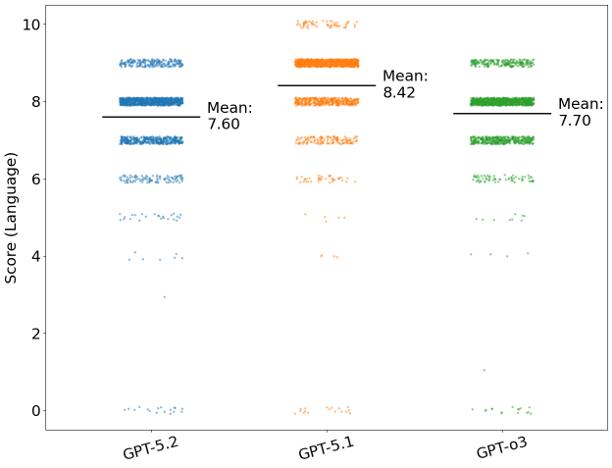}
    \caption{Distribution of language scores.}
    \label{fig:sub3}
  \end{subfigure}

  \caption{Score distributions and mean scores assigned by GPT-5.2, GPT-5.1, and GPT-o3 across three evaluation dimensions. Each dot denotes an individual resume, and the horizontal line indicates the mean score for each  LLM model.}
  \label{fig:gpts_score_distributions}
\end{figure*}

As mentioned above, different recruiters may provide inconsistent judgments for the same resume. 
To simulate this phenomenon, we use multiple LLMs judges to construct the ground truth and analyze their agreement. 
Based on the resulting ground truth, We find that the three LLMs produce different final judgments for 39\% of resumes, indicating substantial differences across the judges.
In addition, as specified by the evaluation instructions in Fig. \ref{fig:eval_instruction}, the LLM outputs a final judgment result for each resume along with three dimension scores for content, structure, and language. 
Fig. \ref{fig:gpts_score_distributions} further compares their score distributions and mean scores across the three evaluation dimensions. 
We observe that GPT-5.1 has the highest mean scores across all dimensions and concentrates more scores in the high score range. 
In contrast, GPT-5.2 and GPT-o3 yield lower mean scores, particularly on the content and structure dimensions, suggesting that they are more conservative in their scoring.

These results confirm that the judgment criteria differ across LLMs. 
This is reflected not only in the limited agreement on the final judgment results, but also in the different score distributions and mean scores across dimensions. 
Therefore, using multiple LLM judges as ground truth better reflects the variability in recruiter judgments in real-world screening.

\subsection{Effects of AutoScreen-FW on Evaluation Dimension Scores for Open-Source LLMs}
\begin{figure*}[htb]
  \centering

  \begin{subfigure}[t]{0.32\textwidth}
    \centering
    \includegraphics[width=\linewidth]{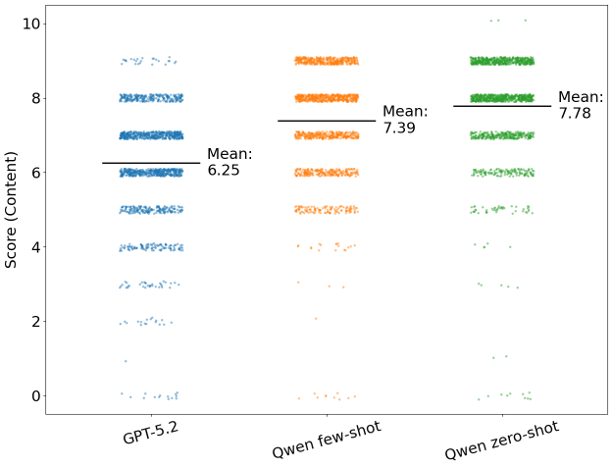}
    \caption{Distribution of content scores.}
    \label{fig:gpt_5.2_content}
  \end{subfigure}\hfill
  \begin{subfigure}[t]{0.32\textwidth}
    \centering
    \includegraphics[width=\linewidth]{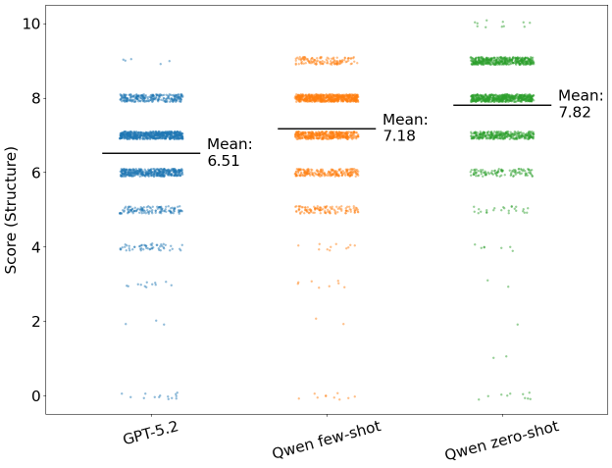}
    \caption{Distribution of structure scores.}
    \label{fig:gpt_5.2_structure}
  \end{subfigure}\hfill
  \begin{subfigure}[t]{0.32\textwidth}
    \centering
    \includegraphics[width=\linewidth]{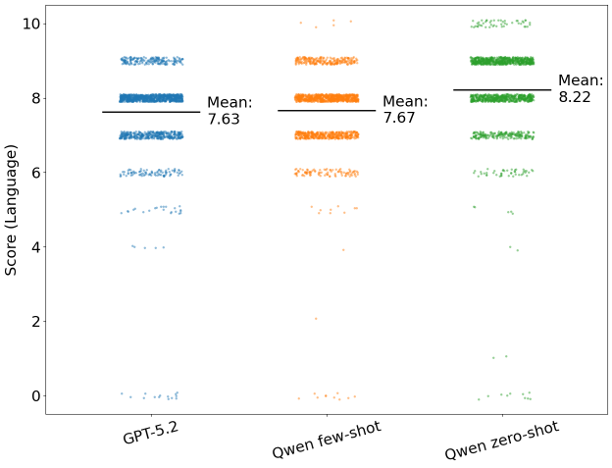}
    \caption{Distribution of language scores.}
    \label{fig:gpt_5.2_language}
  \end{subfigure}

  \caption{Score distributions and mean scores for GPT-5.2, Qwen3-8B (few-shot), and Qwen3-8B (zero-shot) across three evaluation dimensions. Notation follows Fig. \ref{fig:gpts_score_distributions}.}
  \label{fig:gpt5.2_qwen_score}
\end{figure*}

\begin{figure*}[htb]
  \centering
  \begin{subfigure}[t]{0.32\textwidth}
    \centering
    \includegraphics[width=\linewidth]{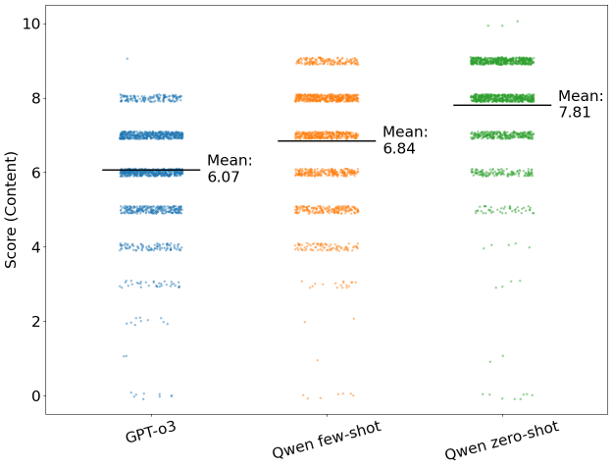}
    \caption{Distribution of content scores.}
    \label{fig:gpt_o3_content}
  \end{subfigure}\hfill
  \begin{subfigure}[t]{0.32\textwidth}
    \centering
    \includegraphics[width=\linewidth]{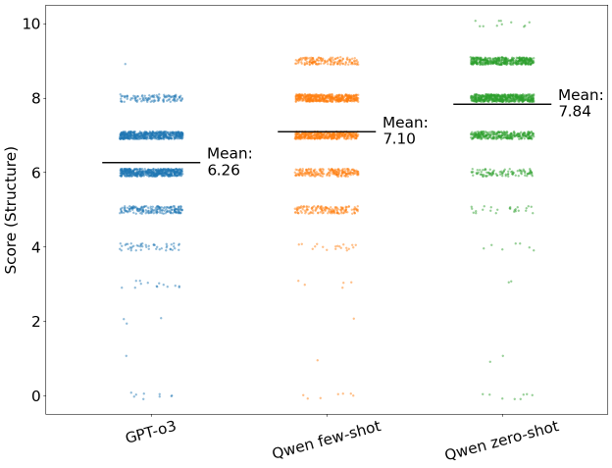}
    \caption{Distribution of structure scores.}
    \label{fig:gpt_o3_structure}
  \end{subfigure}\hfill
  \begin{subfigure}[t]{0.32\textwidth}
    \centering
    \includegraphics[width=\linewidth]{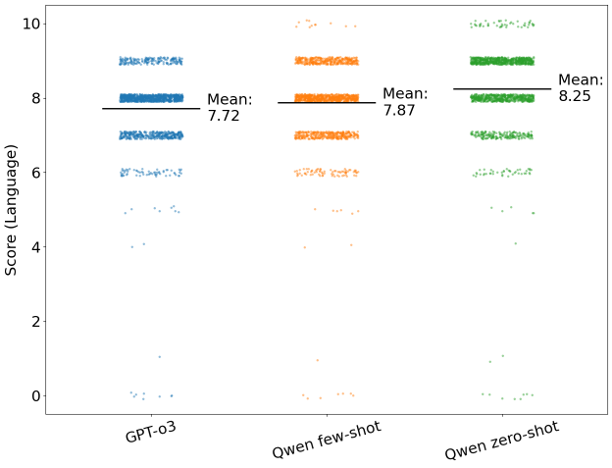}
    \caption{Distribution of language scores.}
    \label{fig:gpt_o3_language}
  \end{subfigure}

  \caption{Score distributions and mean scores for GPT-o3, Qwen3-8B (few-shot), and Qwen3-8B (zero-shot) across three evaluation dimensions. Notation follows Fig. \ref{fig:gpts_score_distributions}.}
  \label{fig:gpto3_qwen_score}
\end{figure*}

Based on Tables \ref{tabl:acc_5.2}, \ref{tabl:acc_5.1}, and \ref{tabl:acc_o3}, we first confirm that under different ground-truth settings, the open-source LLMs Qwen and Llama both improve their judge performance under the AutoScreen-FW. 
In addition, the model outputs dimension scores for each resume.
To analyze how these scores change under different ground truth settings, we compare two ground truths, the newer GPT-5.2 and the relatively older GPT-o3, with Qwen in the zero-shot setting and Qwen with few-shot in-context learning under the AutoScreen-FW. 
We visualize the score distributions and mean scores for the three dimensions in Fig. \ref{fig:gpt5.2_qwen_score} and Fig. \ref{fig:gpto3_qwen_score}.

From the results, we observe two things: (1) under AutoScreen-FW, few-shot Qwen produces dimension score distributions closer to the ground truth than zero-shot Qwen. 
This holds even when the ground truth changes. For example, in Fig. \ref{fig:gpt_o3_content}, zero-shot Qwen assigns content scores mostly between 7 and 9. 
Under few-shot in-context learning, Qwen refers to the evaluation items in Table \ref{tb:es_eval}. 
It also uses GPT-o3's dimension scores for the sample resumes to infer the relationship between resume content and the dimension scores. 
As a result, more resumes receive content scores of 6 or below, and the overall distribution shifts toward GPT-o3. 
In contrast, when GPT-5.2 is used as the ground truth, Qwen does not reference the samples’ dimension scores. 
Even so, it produces dimension score distributions closer to the ground truth by relying only on the samples' final judgments; 
(2) with few-shot in-context learning under AutoScreen-FW, the mean scores across dimensions also move closer to the ground truth, and the mean language score is nearly identical to the ground truth.
Based on these observations, AutoScreen-FW enables Qwen to adjust its scoring behavior to match different judge preferences. 
Moreover, beyond the final judgment for each resume, the three dimension scores produced by Qwen under AutoScreen-FW can also serve as auxiliary information and provide recruiters with more fine-grained guidance. 
This improves the interpretability of resume screening.




\subsection{Efficiency Analysis of LLM-based Resume Screening}
Table \ref{tabl:judge_time} shows the per-resume judgment time for each model. 
Under AutoScreen-FW, Qwen uses only 3 or 5 samples for in-context learning, so per-resume judgment time increases by about 0.1 seconds over the zero-shot setting, which is negligible. 
In contrast, Llama typically uses 15 or 20 samples for in-context learning, leading to a noticeable increase in per-resume time over the zero-shot setting. 
Nevertheless, as discussed in section~\ref{subsec:results}, even with in-context learning, the per-resume judgment time of both open-source LLMs remains substantially lower than that of GPT-5-mini and GPT-5-nano.
In addition, \cite{b38} reports that a recruiter screening of a resume takes 7.4 seconds on average. 
Replacing this step with Qwen can therefore reduce screening time by up to 48\%, increasing screening throughput and easing recruiters’ workload.

\subsection{Visualization of Sampling Strategies}
As discussed in section \ref{subsec:samp_select}, different sampling strategies tend to select samples with different characteristics from the dataset. 
To further examine how the selected samples are distributed in the representation space, we visualize and compare the samples chosen by each strategy. 
Specifically, we encode each resume into a vector using Qwen3-Embedding-8B. 
UMAP\cite{b29} is then applied to project the high-dimensional vectors into a 2D space. 
Here, we use UMAP because it is designed to preserve local neighborhood relationships in the low-dimensional projection, which helps reveal clustering and dispersion patterns in the embedding space. 
We then plot the 2D projections of all resumes and highlight the samples selected by each strategy.

\begin{figure*}[htb]
  \centering
  \begin{subfigure}[t]{0.32\textwidth}
    \centering
    \includegraphics[width=\linewidth]{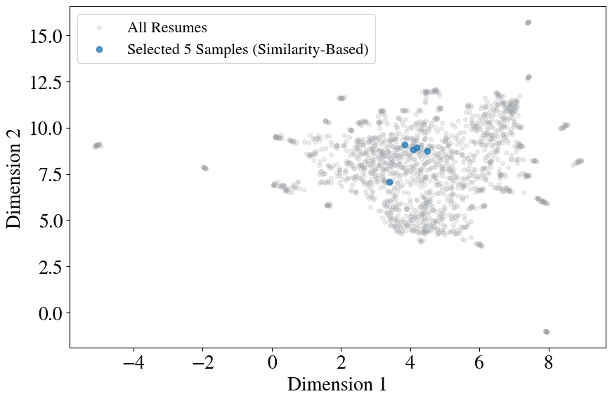}
    \caption{Similarity-based selection.}
    \label{fig:sub1_select_5}
  \end{subfigure}\hfill
  \begin{subfigure}[t]{0.32\textwidth}
    \centering
    \includegraphics[width=\linewidth]{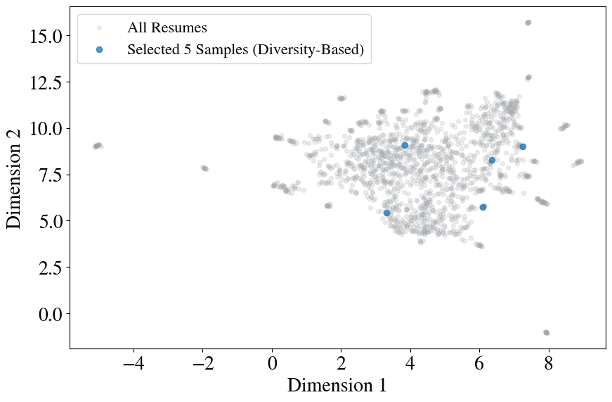}
    \caption{Diversity-based selection.}
    \label{fig:sub2_select_5}
  \end{subfigure}\hfill
  \begin{subfigure}[t]{0.32\textwidth}
    \centering
    \includegraphics[width=\linewidth]{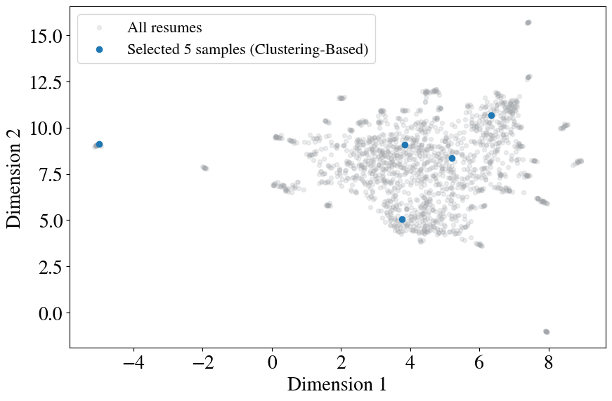}
    \caption{Clustering-based selection.}
    \label{fig:sub3_select_5}
  \end{subfigure}
  \caption{Visualization of resume embeddings highlighting the five samples selected by each sampling strategy.}
  \label{fig:sample_selects_5}
\end{figure*}
\begin{figure*}[htb]
  \centering
  \begin{subfigure}[t]{0.32\textwidth}
    \centering
    \includegraphics[width=\linewidth]{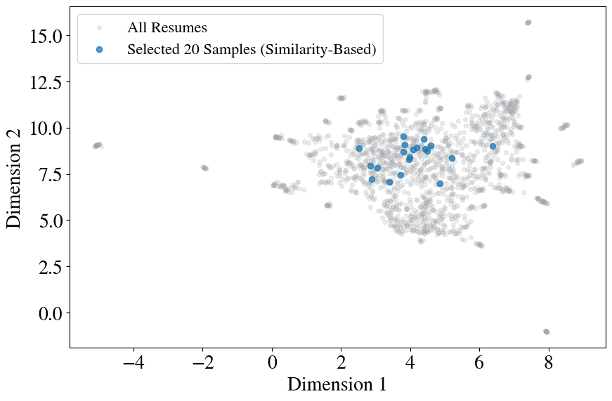}
    \caption{Similarity-based selection.}
    \label{fig:sub1_select_20}
  \end{subfigure}\hfill
  \begin{subfigure}[t]{0.32\textwidth}
    \centering
    \includegraphics[width=\linewidth]{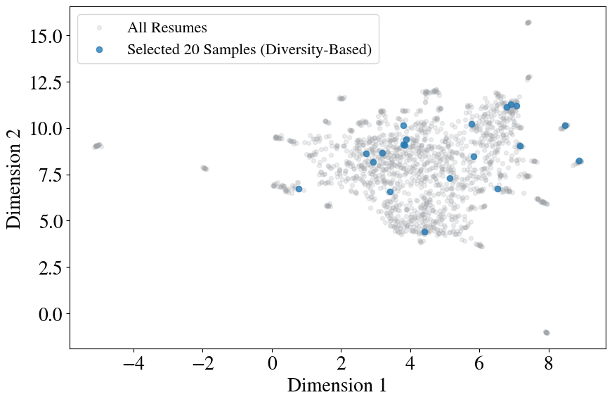}
    \caption{Diversity-based selection.}
    \label{fig:sub2_select_20}
  \end{subfigure}\hfill
  \begin{subfigure}[t]{0.32\textwidth}
    \centering
    \includegraphics[width=\linewidth]{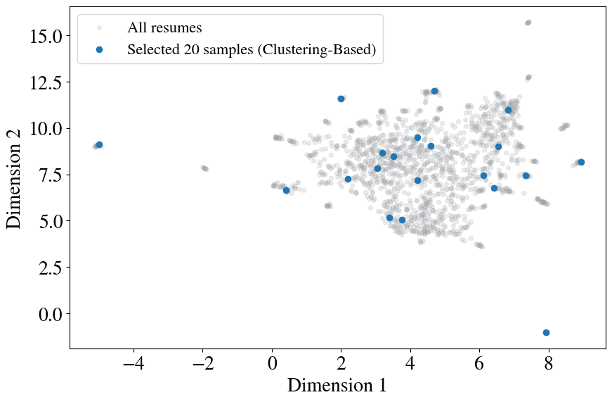}
    \caption{Clustering-based selection.}
    \label{fig:sub3_select_20}
  \end{subfigure}
  \caption{Visualization of resume embeddings highlighting the twenty samples selected by each sampling strategy.}
  \label{fig:sample_selects_20}
\end{figure*}

Fig. \ref{fig:sample_selects_5} and Fig. \ref{fig:sample_selects_20} show the visualization results for selecting five and twenty samples from the dataset with each strategy, respectively.
Fig. \ref{fig:sub1_select_5} and Fig. \ref{fig:sub1_select_20} show that under the similarity-based strategy, the selected samples concentrate near the center of the distribution in both the five-sample and twenty-sample settings. 
This indicates that the strategy tends to choose typical samples that are most similar to the dataset as a whole. 
Fig. \ref{fig:sub2_select_5} and Fig. \ref{fig:sub2_select_20} show that under the diversity-based strategy, the selected samples are farther apart from each other and span a wide area, indicating larger differences in their content-related features and thus higher diversity.
Fig. \ref{fig:sub3_select_5} and Fig. \ref{fig:sub3_select_20} further show that under the clustering-based strategy, the selected samples also achieve broad coverage. 
Moreover, compared with the diversity-based strategy, this strategy selects more samples near the boundary of the distribution. 
This suggests that the clustering-based strategy covers more extreme samples, leading to higher diversity.


Finally, based on the results in Tables \ref{tabl:acc_5.2}, \ref{tabl:acc_5.1}, \ref{tabl:acc_o3}, we observe that Llama generally achieves better performance when using more diverse samples. 
In contrast, Qwen’s preferred sample characteristics vary across different ground-truth settings, suggesting that it requires example selection that better matches the target ground truth to achieve the best performance.

\section{Limitation}
Based on the results in Tables \ref{tabl:acc_5.2}, \ref{tabl:acc_5.1}, \ref{tabl:acc_o3}, to achieve the highest judgment accuracy under different ground truth settings, the reference sample setup must be adjusted, including the sampling strategy, the number of examples, the sample type, and the attribute type. 
Although AutoScreen-FW supports flexible combinations of these options, the optimal configuration still requires manual tuning, and the framework cannot automatically select the best parameter setting when the ground truth changes. 
This limitation leads to additional tuning when the evaluation criteria change. 
Moreover, our experiments rely on LLM-constructed ground truth. 
While this setup can partially capture variability across evaluators, it may still differ from the standards used by real corporate recruiters. 
Therefore, the effectiveness of AutoScreen-FW in real-world hiring requires further validation.

\section{Conclusion}
In this paper, we propose AutoScreen-FW, a framework for a framework for locally and automatically screening resumes. 
The framework defines the key components for LLM-based judging, including sample selection method, persona descriptions for the LLM, resume evaluation criteria, and evaluation instructions. 
We conduct experiments using three high-performing GPT models as ground truth. 
The experiment results show that, with few-shot in-context learning under AutoScreen-FW, the judge performance of open-source LLMs consistently surpasses GPT-5-nano under all three ground-truth settings. 
Moreover, when using GPT-5.1 as the ground truth, the performance further exceeds GPT-5-mini. 
Although performance remains slightly below GPT-5-mini under the other ground-truth settings, the per-resume judgment time can be reduced by 27\%. 
Based on these results, AutoScreen-FW can be deployed on-premise in companies for resume screening, reducing recruiters' workload and allowing them to focus on higher-value tasks.

Future work will proceed in two directions. 
First, we will incorporate an automated sample selection module that chooses the most suitable sampling strategy as well as the sample types and attributes for each ground-truth setting. 
Second, we will further validate the agreement between the resume evaluations produced by open-source LLMs under AutoScreen-FW and those provided by corporate recruiters.


\end{document}